\documentclass[11pt]{article}
\usepackage[T2A,T1]{fontenc}
\usepackage[utf8]{inputenc}
\usepackage[russian,english]{babel}
\usepackage{nodalida2019}
\usepackage{graphicx}
\usepackage{times}
\usepackage{url}
\usepackage{hyperref}
\usepackage{latexsym}
\usepackage{booktabs}
\usepackage{etoolbox,refcount}
\usepackage{multicol}

\newcounter{countitems}
\newcounter{nextitemizecount}
\newcommand{\setupcountitems}{%
  \stepcounter{nextitemizecount}%
  \setcounter{countitems}{0}%
  \preto\item{\stepcounter{countitems}}%
}
\makeatletter
\newcommand{\computecountitems}{%
  \edef\@currentlabel{\number\c@countitems}%
  \label{countitems@\number\numexpr\value{nextitemizecount}-1\relax}%
}
\newcommand{\nextitemizecount}{%
  \getrefnumber{countitems@\number\c@nextitemizecount}%
}
\newcommand{\previtemizecount}{%
  \getrefnumber{countitems@\number\numexpr\value{nextitemizecount}-1\relax}%
}
\makeatother    
{\end{itemize}%
\unskip\computecountitems\ifnumcomp{\previtemizecount}{>}{3}{\end{multicols}}{}}

\aclfinalcopy 
\def\aclpaperid{3} 

\title{To lemmatize or not to lemmatize: how word normalisation affects ELMo performance in word sense disambiguation} 

\author{Andrey Kutuzov\thanks{Both authors contributed equally to the paper.}\\
  University of Oslo \\
  Oslo, Norway \\
  {\tt andreku@ifi.uio.no} \\\And
  Elizaveta Kuzmenko \\
  University of Trento \\
  Trento, Italy  \\
  {\tt lizaku77@gmail.com}
  }

\date{}

\begin{document}
\maketitle
\begin{abstract}
In this paper, we critically evaluate the widespread assumption that deep learning NLP models do not require lemmatized input. To test this, we trained versions of contextualised word embedding \textit{ELMo} models on raw tokenized corpora and on the corpora with word tokens replaced by their lemmas. Then, these models were evaluated on the word sense disambiguation task. This was done for the English and Russian languages.

The experiments showed that while lemmatization is indeed not necessary for English, the situation is different for Russian. It seems that for rich-morphology languages, using lemmatized training and testing data yields small but consistent improvements: at least for word sense disambiguation. This means that the decisions about text pre-processing before training \textit{ELMo} should consider the linguistic nature of the language in question.
\end{abstract}

\section{Introduction} \label{sec:intro}
Deep contextualised representations of linguistic entities (words and/or sentences) are used in many current state-of-the-art NLP systems. The most well-known examples of such models are arguably \textit{ELMo} \cite{peters-etal-2018-deep} and \textit{BERT} \cite{devlin-etal-2019-bert}. 

A long-standing tradition if the field of applying deep learning to NLP tasks can be summarised as follows: as minimal pre-processing as possible. It is widely believed that lemmatization or other text input normalisation is not necessary. Advanced neural architectures based on character input (CNNs, BPE, etc) are supposed to be able to learn how to handle spelling and morphology variations themselves, even for languages with rich morphology: `just add more layers!'. Contextualised embedding models follow this tradition: as a rule, they are trained on raw text collections, with minimal linguistic pre-processing. Below, we show that this is not entirely true.

It is known that for the previous generation of word embedding models (`static' ones like \textit{word2vec} \cite{Mikolov:2013}, where a word always has the same representation regardless of the context in which it occurs), lemmatization of the training and testing data improves their performance. \citet{fares-etal-2017-word} showed that this is true at least for semantic similarity and analogy tasks.

In this paper, we describe our experiments in finding out whether lemmatization helps modern contextualised embeddings (on the example of \textit{ELMo}). We compare the performance of \textit{ELMo} models trained on the same corpus before and after lemmatization. It is impossible to evaluate contextualised models on `static' tasks like lexical semantic similarity or word analogies. Because of this, we turned to \textbf{word sense disambiguation in context} (WSD) as an evaluation task.

In brief, we use contextualised representations of ambiguous words from the top layer of an  \textit{ELMo} model to train word sense classifiers and find out whether using lemmas instead of tokens helps in this task (see Section \ref{sec:experiments}). We experiment with the English and Russian languages and show that they differ significantly in the influence of lemmatization on the WSD performance of ELMo models. 

Our findings and the contributions of this paper are:
\begin{enumerate}
    \item Linguistic text pre-processing still matters in some tasks, even for contemporary deep representation learning algorithms.
    \item For the Russian language, with its rich morphology, lemmatizing the training and testing data for \textit{ELMo} representations yields small but consistent improvements in the WSD task. This is unlike English, where the differences are negligible.
\end{enumerate}

\begin{table}
\center
\begin{tabular}{lcc}
\toprule
&\textbf{English}&\textbf{Russian} \\
\midrule
\textbf{Source} & Wikipedia & Wikipedia + RNC \\
\textbf{Size, tokens} & 2 174 mln & 989 mln  \\
\textbf{Size, lemmas} & 1 977 mln & 988 mln \\
\bottomrule
\end{tabular}
\caption{Training corpora}
\label{tab:corpora}
\end{table}

\section{Related work} \label{sec:related}

\textit{ELMo} contextual word representations are learned in an unsupervised way through language modelling \cite{peters-etal-2018-deep}.  The general architecture consists of a two-layer BiLSTM on top of a convolutional layer which takes character sequences as its input. Since the model uses fully character-based token representations, it avoids the problem of out-of-vocabulary words. Because of this, the authors explicitly recommend not to use any normalisation except tokenization for the input text. However, as we show below, while this is true for English, for other languages feeding \textit{ELMo} with lemmas instead of raw tokens can improve WSD performance.

Word sense disambiguation or WSD \cite{navigli2009word} is the NLP task consisting of choosing a word sense from a pre-defined sense inventory, given the context in which the word is used. WSD fits well into our aim to intrinsically evaluate \textit{ELMo} models, since solving the problem of polysemy and homonymy was one of the original promises of contextualised embeddings: their primary difference from the previous generation of word embedding models is that contextualised approaches generate different representations for homographs depending on the context. We use two lexical sample WSD test sets, further described in Section \ref{sec:datasets}.

\section{Training ELMo} \label{sec:elmo}

For the experiments described below, we trained our own \textit{ELMo} models from scratch. For English, the training corpus consisted of the English Wikipedia dump\footnote{\url{https://dumps.wikimedia.org/}} from February 2017. For Russian, it was a concatenation of the Russian Wikipedia dump from December 2018 and the full Russian National Corpus\footnote{\url{http://ruscorpora.ru/en/}} (RNC). The RNC texts were added to the Russian Wikipedia dump so as to make the Russian training corpus more comparable in size to the English one (Wikipedia texts would comprise only half of the size). As Table \ref{tab:corpora} shows, the English Wikipedia is still two times larger, but at least the order is the same.

The texts were tokenized and lemmatized with the \textit{UDPipe} models for the respective languages trained on the Universal Dependencies 2.3 treebanks \cite{straka-strakova-2017-tokenizing}. \textit{UDPipe} yields lemmatization accuracy about 96\% for English and 97\% for Russian\footnote{\url{http://ufal.mff.cuni.cz/udpipe/models\#universal\_dependencies\_23\_models}}; thus for the task at hand, we considered it to be gold and did not try to further improve the quality of normalisation itself (although it is not entirely error-free, see Section \ref{sec:datasets}).

\textit{ELMo} models were trained on these corpora using the original TensorFlow implementation\footnote{\url{https://github.com/allenai/bilm-tf}}, for 3 epochs with batch size 192, on two GPUs. To train faster, we decreased the dimensionality of the LSTM layers from the default 4096 to 2048 for all the models.

\section{Word sense disambiguation test sets} \label{sec:datasets}

We used two WSD datasets for evaluation:
\begin{itemize}
    \item \textit{Senseval-3} for English \cite{mihalcea-etal-2004-senseval}
    \item \textit{RUSSE'18} for Russian \cite{Panchenko:18:dialogue}
\end{itemize}
  
The \textit{Senseval-3} dataset consists of lexical samples for nouns, verbs and adjectives; we used only noun target words:
\begin{enumerate}
    \item \textit{argument}
    \item \textit{arm}
    \item \textit{atmosphere}
    \item \textit{audience}
    \item \textit{bank}
    \item \textit{degree}
    \item \textit{difference}
    \item \textit{difficulty}
    \item \textit{disc}
    \item \textit{image}
    \item \textit{interest}
    \item \textit{judgement}
    \item \textit{organization}
    \item \textit{paper}
    \item \textit{party}
    \item \textit{performance}
    \item \textit{plan}
    \item \textit{shelter}
    \item \textit{sort}
    \item \textit{source}
\end{enumerate}

An example for the ambiguous word \textit{argument} is given below: 
{\par \textit{In some situations Postscript can be faster than the escape sequence type of printer control file. It uses post fix notation, where \textbf{arguments} come first and operators follow. This is basically the same as Reverse Polish Notation as used on certain calculators, and follows directly from the stack based approach.} }

It this sentence, the word `\textit{argument}' is used in the sense of a mathematical operator.

The \textit{RUSSE'18} dataset was created in 2018 for the shared task in Russian word sense induction. This dataset contains only nouns; the list of words with their English translations is given in Table \ref{tab:rus}.

\begin{table}
\center
\begin{tabular}{ll}
\toprule
\textbf{Target word} & \textbf{Translation} \\
\midrule
\foreignlanguage{russian}{акция} & `stock/marketing event' \\
\foreignlanguage{russian}{гипербола} & `hyperbola/exaggeration' \\
\foreignlanguage{russian}{град} & `hail/city' \\
\foreignlanguage{russian}{гусеница} & `caterpillar/track' \\
\foreignlanguage{russian}{домино} & `dominoes/costume' \\
\foreignlanguage{russian}{кабачок} & `squash/restaurant' \\
\foreignlanguage{russian}{капот} & `hood (part of a car/clothing)' \\
\foreignlanguage{russian}{карьер} & `mine/fast pace of a horse' \\
\foreignlanguage{russian}{кок} & `cook/hairstyle' \\
\foreignlanguage{russian}{крона} & `crown (tree/coin)' \\
\foreignlanguage{russian}{круп} & `crupper (part of a horse/illness)' \\
\foreignlanguage{russian}{мандарин} & `fruit/a Chinese official' \\
\foreignlanguage{russian}{рок} & `rock (music/destiny)' \\
\foreignlanguage{russian}{слог} & `syllable/text style' \\
\foreignlanguage{russian}{стопка} & `stack/glass' \\
\foreignlanguage{russian}{таз} & `basin/human body part' \\
\foreignlanguage{russian}{такса} & `tariff/dog breed' \\
\foreignlanguage{russian}{шах} & `check/prince' \\
\bottomrule
\end{tabular}
\caption{Target ambiguous words for Russian (\textit{RUSSE'18})}
\label{tab:rus}
\end{table}

Originally, it includes also the words \textit{\foreignlanguage{russian}{байка}} `tale/fleece' and \textit{\foreignlanguage{russian}{гвоздика}} 'clove/small nail', but their senses are ambiguous only in some inflectional forms (not in lemmas), therefore we decided to exclude these words from evaluation.

The Russian dataset is more homogeneous compared to the English one, as for all the target words there is approximately the same number of context words in the examples. This is achieved by applying the lexical window (25 words before and after the target word) and cropping everything that falls outside of that window. In the English dataset, on the contrary, the whole paragraph with the target word is taken into account. We have tried cropping the examples for English as well, but it did not result in any change in the quality of classification. In the end, we decided not to apply the lexical window to the English dataset so as not to alter it and rather use it in the original form. 

Here is an example from the \textit{RUSSE'18} for the ambiguous word \textit{\foreignlanguage{russian}{мандарин}} `mandarin' in the sense `Chinese official title': {\par \textit{\foreignlanguage{russian}{``...дипломатического корпуса останкам богдыхана и императрицы обставлено было с необычайной торжественностью. Тысячи мандаринов и других высокопоставленных лиц разместились шпалерами на трех мраморных террасах ведущих к...''}} \par`...the diplomatic bodies of the Bogdikhan and the Empress was furnished with extraordinary solemnity. Thousands of mandarins and other dignitaries were placed on three marble terraces leading to...'.}

Table \ref{tab:datasets} compares both datasets. Before usage, they were pre-processed in the same way as the training corpora for \textit{ELMo} (see Section \ref{sec:elmo}), thus producing a lemmatized and a non-lemmatized versions of each.

\begin{table}
\center
\begin{tabular}{lcc}
\toprule
\textbf{Property} & \textbf{Senseval-3} & \textbf{RUSSE'18} \\
\midrule
\textbf{Target words} & 20 & 18 \\
\textbf{Distinct target forms} & 39 & 132 \\
\textbf{Distinct target lemmas} & 24 & 36 \\
\midrule
\textbf{Examples per target} & 171 & 126 \\
\textbf{Tokens per example} & 126 & 25  \\
\textbf{Senses per target} & 6 & 2 \\
\bottomrule
\end{tabular}
\caption{Characteristics of the WSD datasets. The numbers in the lower part are average values.}
\label{tab:datasets}
\end{table}

As we can see from Table \ref{tab:datasets}, for 20 target words in English there are 24 lemmas, and for 18 target words in Russian there are 36 different lemmas. These numbers are explained by occasional errors in the \textit{UDPipe} lemmatization. 
Another interesting thing to observe is the number of distinct word forms for every language. For English, there are 39 distinct forms for 20 target nouns: singular and plural for every noun, except `\textit{atmosphere}' which is used only in the singular form. Thus, inflectional variability of English nouns is covered by the dataset almost completely. For Russian, we observe 132 distinct forms for 18 target nouns, giving more than 7 inflectional forms per each word. Note that this still covers only half of all the inflectional variability of Russian: this language features 12 distinct forms for each noun (6 cases and 2 numbers). 

To sum up, the \textit{RUSSE'18} dataset is morphologically far more complex than the \textit{Senseval3}, reflecting the properties of the respective languages. In the next section we will see that this leads to substantial differences regarding comparisons between token-based and lemma-based \textit{ELMo} models.

\section{Experiments}\label{sec:experiments}
Following \citet{gorman-bedrick-2019-need}, we decided to avoid using any standard train-test splits for our WSD datasets. Instead, we rely on per-word random splits and 5-fold cross-validation. This means that for each target word we randomly generate 5 different divisions of its context sentences list into train and test sets, and then train and test 5 different classifier models on this data. The resulting performance score for each target word is the average of 5 macro-F1 scores produced by these classifiers.

\textit{ELMo} models can be employed for the WSD task in two different ways: either by fine-tuning the model or by extracting word representations from it and then using them as features in a downstream classifier. We decided to stick to the second (feature extraction) approach, since it is conceptually and computationally simpler. Additionally, \citet{peters-etal-2019-tune} showed that for most NLP tasks (except those focused on sentence pairs) the performance of feature extraction and fine-tuning is nearly the same. Thus we extracted the single vector of the target word from the \textit{ELMo} top layer (`target' rows in Table \ref{tab:scores}) or the averaged \textit{ELMo} top layer vectors of all words in the context sentence (`averaged' rows in Table \ref{tab:scores}). 

\begin{table}
\center
\begin{tabular}{lcc}
\toprule
\textbf{Model}&\textbf{English}&\textbf{Russian} \\
\midrule
& \multicolumn{2}{c}{Baselines} \\
\midrule
Random & $\approx0.138$ & $\approx0.444$ \\
MFS & 0.119 & 0.391 \\
\midrule
& \multicolumn{2}{c}{Tokens} \\
\midrule
SGNS (averaged) & 0.299  & 0.851  \\
\textit{ELMo} (averaged) & 0.362 & 0.885 \\
\textit{ELMo} (target) & 0.463 & 0.875 \\
\midrule
& \multicolumn{2}{c}{Lemmas} \\
\midrule
SGNS (averaged) & 0.300 & 0.854 \\
\textit{ELMo} (averaged) & 0.365 & 0.888  \\
\textit{ELMo} (target) & 0.452 & \textbf{0.907} \\
\bottomrule
\end{tabular}
\caption{Averaged macro-F1 scores for WSD}
\label{tab:scores}
\end{table}

For comparison, we also report the scores of the  `averaged vectors' representations with Continuous Skipgram \cite{Mikolov:2013} embedding models trained on the English or Russian Wikipedia dumps (`SGNS' rows): before the advent of contextualised models, this was one of the most widely used ways to `squeeze' the meaning of a sentence into a fixed-size vector. Of course it does not mean that the meaning of a sentence always determines the senses all its words are used in. However, averaging representations of words in contexts as a proxy to the sense of one particular word is a long established tradition in WSD, starting at least from \citet{schutze-1998-automatic}. Also, since SGNS is a `static' embedding model, it is of course not possible to use only target word vectors as features: they would be identical whatever the context is.

Simple logistic regression was used as a classification algorithm. We also tested a multi-layer perceptron (MLP) classifier with 200-neurons hidden layer, which yielded essentially the same results. This leads us to believe that our findings are not classifier-dependent.

Table \ref{tab:scores} shows the results, together with the random and most frequent sense (MFS) baselines for each dataset.

First, \textit{ELMo} outperforms SGNS for both languages, which comes as no surprise. Second, the approach with averaging representations from all words in the sentence is not beneficial for WSD with \textit{ELMo}: for English data, it clearly loses to a single target word representation, and for Russian there are no significant differences (and using a single target word is preferable from the computational point of view, since it does not require the averaging operation). Thus, below we discuss only the single target word usage mode of \textit{ELMo}.

\begin{figure}
    \centering
    \includegraphics[scale=0.5,keepaspectratio]{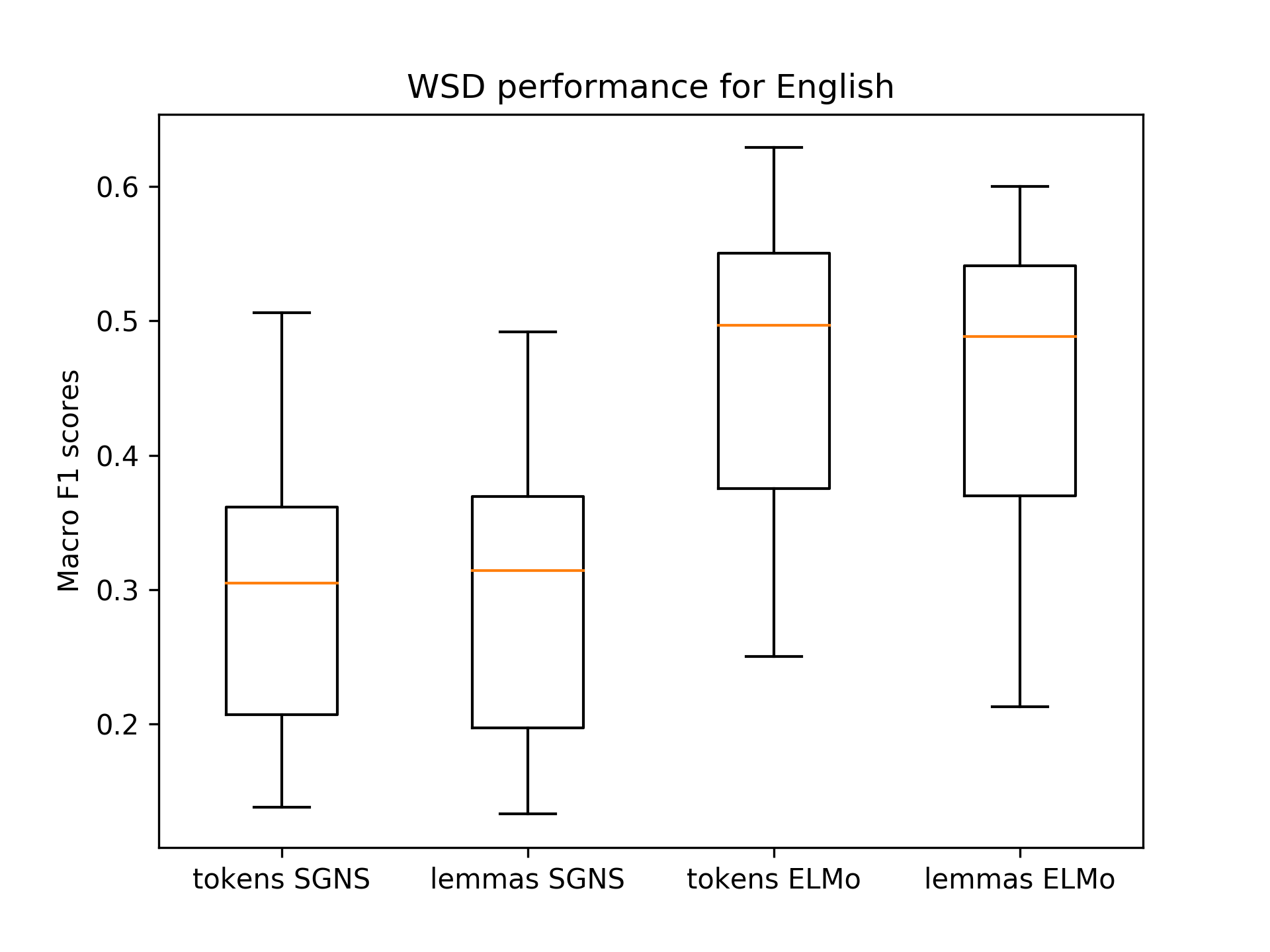}
    \caption{Word sense disambiguation performance on the \textbf{English} data across words (\textit{ELMo} target models).}
    \label{fig:english}
\end{figure}

\begin{figure}
    \centering
    \includegraphics[scale=0.5,keepaspectratio]{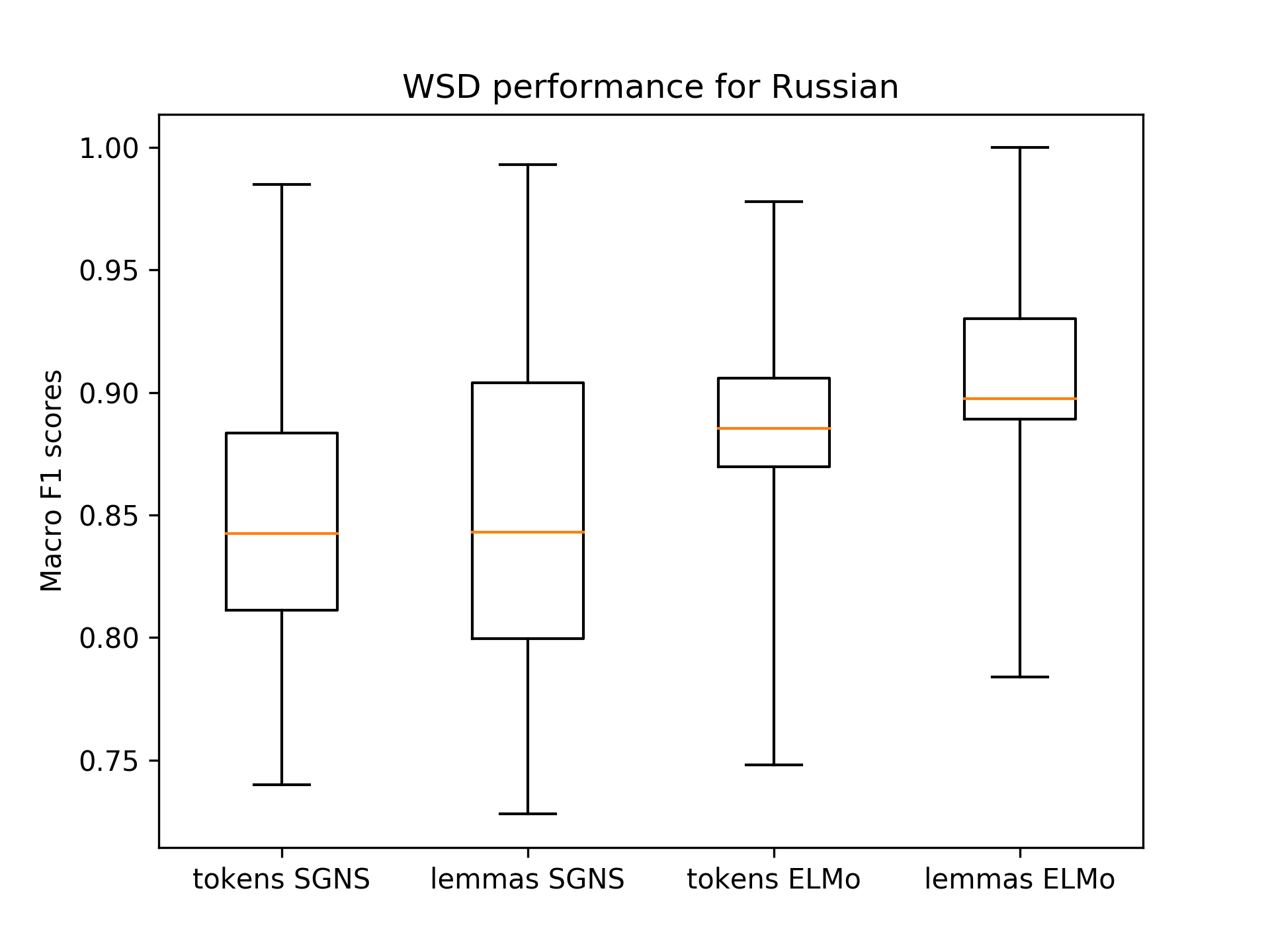}
    \caption{Word sense disambiguation performance on the \textbf{Russian} data across words (\textit{ELMo} target models).}
    \label{fig:russian}
\end{figure}

But the most important part is the comparison between using tokens or lemmas in the train and test data. For the `static' SGNS embeddings, it does not significantly change the WSD scores for both languages. The same is true for English \textit{ELMo} models, where differences are negligible and seem to be simple fluctuations. However, for Russian, \textit{ELMo} (target) on lemmas outperforms \textit{ELMo} on tokens, with small but significant\footnote{At $p$ value of 0.1, according to the Welch's t-test.} improvement. The most plausible explanation for this is that (despite of purely character-based input of \textit{ELMo}) the model does not have to learn idiosyncrasies of a particular language morphology. Instead, it can use its (limited) capacity to better learn lexical semantic structures, leading to better WSD performance. The box plots \ref{fig:english} and \ref{fig:russian} illustrate the scores dispersion across words in the test sets for English and Russian correspondingly (orange lines are medians).  In the next section \ref{sec:analysis} we analyse the results qualitatively.

\section{Qualitative analysis} \label{sec:analysis}

In this section we focus on the comparison of scores for the Russian dataset. 
The classifier for Russian had to choose between fewer classes (two or three), which made the scores higher and more consistent than for the English dataset. Overall, we see improvements in the scores for the majority of words, which proves that lemmatization for morphologically rich languages is beneficial.

We decided to analyse more closely those words for which the difference in the scores between lemma-based and token-based models was statistically significant. By `significant' we mean that the scores differ by more that one standard deviation (the largest standard deviation value in the two sets was taken). The resulting list of targets words with significant difference in scores is given in Table \ref{tab:comp_russian}.

\begin{table}
\center
\begin{tabular}{lccc}
\toprule
\textbf{Word} & \textbf{Tokens} & \textbf{Lemmas} & \textbf{STD} \\
\midrule
\textbf{\foreignlanguage{russian}{акция}} & 0.876 &  \textbf{0.978} & 0.050 \\
\textbf{\foreignlanguage{russian}{крона}} & 0.978 & \textbf{1.000} & 0.018 \\
\textbf{\foreignlanguage{russian}{круп}} & 0.927 & \textbf{1.000}  & 0.070 \\
\midrule
\textbf{\foreignlanguage{russian}{домино}}  & \textbf{0.910} & 0.874 & 0.057 \\
\bottomrule
\end{tabular}
\caption{F1 scores for target words from \textit{RUSSE'18} with significant differences between lemma-based and token-based models}
\label{tab:comp_russian}
\end{table}

We can see that among 18 words in the dataset only 3 exhibit significant improvement in their scores when moving from tokens to lemmas in the input data. It shows that even though the overall F1 scores for the Russian data have shown the plausibility of lemmatization, this improvement is mostly driven by a few words. 
It should be noted that these words' scores feature very low standard deviation values (for other words, standard deviation values were above 0.1, making F1 differences insignificant). Such a behaviour can be caused by more consistent differentiation of context for various senses of these 3 words. For example, with the word \foreignlanguage{russian}{кабачок} `squash / small restaurant', the contexts for both senses can be similar, since they are all related to food. This makes the WSD scores unstable. On the other hand, for \foreignlanguage{russian}{акция} `stock, share / event', \foreignlanguage{russian}{крона} `crown (tree / coin)' or \foreignlanguage{russian}{круп} `croup (horse body part / illness)', their senses are not related, which resulted in more stable results and significant difference in the scores (see Table \ref{tab:comp_russian}).

There is only one word in the RUSSE'18 dataset for which the score has strongly decreased when moving to lemma-based models: \foreignlanguage{russian}{домино} `domino (game / costume)'. In fact, the score difference here lies on the border of one standard deviation, so strictly speaking it is not really significant. However, the word still presents an interesting phenomenon. 

\foreignlanguage{russian}{Домино} is the only target noun in the RUSSE'18 that has no inflected forms, since it is a borrowed word. This leaves no room for improvement when using lemma-based \textit{ELMo} models: all tokens of this word are already identical. At the same time, some information about inflected word forms in the context can be useful, but it is lost during lemmatization, and this leads to the decreased score. Arguably, this means that lemmatization brings along both advantages and disadvantages for WSD with \textit{ELMo}. For inflected words (which constitute the majority of Russian vocabulary) profits outweigh the losses, but for atypical non-changeable words it can be the opposite.  

The scores for the excluded target words \textit{\foreignlanguage{russian}{байка}} `tale / fleece' and \textit{\foreignlanguage{russian}{гвоздика}} 'clove / small nail' are given in Table \ref{tab:comp_excluded} (recall that they were excluded because of being ambiguous only in some inflectional forms). For these words we can see a great improvement with lemma-based models. This, of course stems from the fact that these words in different senses have different lemmas. Therefore, the results are heavily dependent on the quality of lemmatization.

\begin{table}
\center
\begin{tabular}{lccc}
\toprule
\textbf{Word} & \textbf{Tokens} & \textbf{Lemmas} & \textbf{STD} \\
\midrule
\textit{\foreignlanguage{russian}{байка}} & 0.421 & \textbf{0.627}  & 0.099 \\ 
\textit{\foreignlanguage{russian}{гвоздика}} & 0.553 & \textbf{0.619} & 0.038 \\
\bottomrule
\end{tabular}
\caption{F1 scores for the excluded target words from \textit{RUSSE'18}.}
\label{tab:comp_excluded}
\end{table}

\section{Conclusion} \label{sec:conclusion}

We evaluated how the ability of \textit{ELMo} contextualised word embedding models to disambiguate word senses depends on the nature of the training data. In particular, we compared the models trained on raw tokenized corpora and those trained on the corpora with word tokens replaced by their normal forms (lemmas). The models we trained are publicly available via the NLPL word embeddings repository\footnote{\url{http://vectors.nlpl.eu/repository/}} \cite{fares-etal-2017-word}. 

In the majority of research papers on deep learning approaches to NLP, it is assumed that lemmatization is not necessary, especially when using powerful contextualised embeddings. Our experiments show that this is indeed true for languages with simple morphology (like English). However, for rich-morphology languages (like Russian), using lemmatized training data yields small but consistent improvements in the word sense disambiguation task. These improvements are not observed for rare words which lack inflected forms; this further supports our hypothesis that better WSD scores of lemma-based models are related to them better handling multiple word forms in morphology-rich languages. 

Of course, lemmatization is by all means not a silver bullet. In other tasks, where inflectional properties of words are important, it can even hurt the performance. But this is true for any NLP systems, not only deep learning based ones.

The take-home message here is twofold: first, text pre-processing still matters for contemporary deep learning algorithms. Their impressive learning abilities do not always allow them to infer normalisation rules themselves, from simply optimising the language modelling task. Second, the nature of language at hand matters as well, and differences in this nature can result in different decisions being optimal or sub-optimal at the stage of deep learning models training. The simple truth `English is not representative of all languages on Earth' still holds here.

In the future, we plan to extend our work by including more languages into the analysis. Using Russian and English allowed us to hypothesise about the importance of morphological character of a language. But we only scratched the surface of the linguistic diversity. To verify this claim, it is necessary to analyse more strongly inflected languages like Russian as well as more weakly inflected (analytical) languages similar to English. This will help to find out if the inflection differences are important for training deep learning models across human languages in general.

\bibliographystyle{acl_natbib}
\bibliography{nodalida2019}

\end{document}